\pdfoutput=1

\documentclass[11pt]{article}

\usepackage[preprint]{acl}

\usepackage{times}
\usepackage{latexsym}

\usepackage[T1]{fontenc}

\usepackage[utf8]{inputenc}

\usepackage{microtype}

\usepackage{inconsolata}

\usepackage{graphicx}

\title{Guiding and Diversifying LLM-Based Story Generation\\via Answer Set Programming}

\author{
Phoebe J. Wang\textsuperscript{1} \and Max Kreminski\textsuperscript{1,2}
\\
\textsuperscript{1}Santa Clara University,
\textsuperscript{2}Midjourney
\\
\texttt{wjingsen@gmail.com}, \texttt{mkreminski@midjourney.com}
}

\begin{document}
\maketitle
\begin{abstract}
Instruction-tuned large language models (LLMs) are capable of generating stories in response to open-ended user requests, but the resulting stories tend to be limited in their diversity. Older, symbolic approaches to story generation (such as planning) can generate substantially more diverse plot outlines, but are limited to producing stories that recombine a fixed set of hand-engineered character action templates. Can we combine the strengths of these approaches while mitigating their weaknesses? We propose to do so by using a higher-level and more abstract symbolic specification of high-level story structure---implemented via answer set programming (ASP)---to guide and diversify LLM-based story generation. Via semantic similarity analysis, we demonstrate that our approach produces more diverse stories than an unguided LLM, and via code excerpts, we demonstrate the improved compactness and flexibility of ASP-based outline generation over full-fledged narrative planning.
\end{abstract}

\section{Introduction}
Story generation research today has two major prongs: one that aims to use symbolic or rules-based AI methods (such as planning) to generate plot outlines via hand-crafted bases of story structure knowledge, and one that aims to produce fully realized textual stories via large language models (LLMs). Both approaches have significant strengths and weaknesses. Planners are good at reasoning about long-term story structure and ensuring plot coherence~\cite{BalancingPlotAndCharacter,NarrativePlanning}, but narrative planning domains are notoriously effortful to hand-engineer~\cite{NarrativeActionSchemasSuspense,Siler}, and their fine-grained encoding of possible character actions and plot elements make them rigid in their adherence to a particular, narrow \emph{type} of story.

LLMs, on the other hand, are effective at open-domain story generation. 
However, they tend to struggle with long-term story coherence, often failing to produce a satisfying overall story arc or reader experience~\cite{UnmetNeeds}. Additionally, comparing LLM-generated to human-written stories reveals many aesthetic weaknesses~\cite{ArtOrArtifice,Dramatron}, in particular a tendency toward unoriginality or \emph{homogeneity} in both the semantic and syntactic aspects of generated stories~\cite{HomogenizationEffects,Begus}.

The clear complementarity of symbolic and neural approaches to story generation has led many researchers to try knitting these approaches together in various ways~\cite{Martin,TattleTale,ThereAndBackAgain,StoryVerse}. Nevertheless, the rigidity of narrative planning has made it difficult to apply planners to open-domain story generation, even as ``plan-and-write'' approaches to story generation~\cite{PlanAndWrite} have grown in popularity on the LLM side.

In this paper, we propose a slight variation on the plan-and-write approach---i.e., first generating a story outline and then expanding that outline into a complete story via LLM. Unlike past approaches, we use neither a language model~\cite{HNSG,PlanAndWrite,AristotelianContentPlanning} nor a fine-grained symbolic planning domain~\cite{ThereAndBackAgain,StoryVerse} to generate story outlines. Instead, we use a lightweight microtheory of storytelling goal sequencing encoded as an \emph{answer set program}. This enables us to combine the open-domain storytelling capabilities of an LLM 
with the high-level structural and diversity benefits of a symbolic approach to outline generation. 
We demonstrate the potential compactness and flexibility of an ASP-based outline generator through code excerpts, and the diversity benefits of our approach through a quantitative analysis of semantic similarities between sets of generated stories.

\section{Our Approach}
\begin{figure*}
    \centering
    \includegraphics[width=\linewidth]{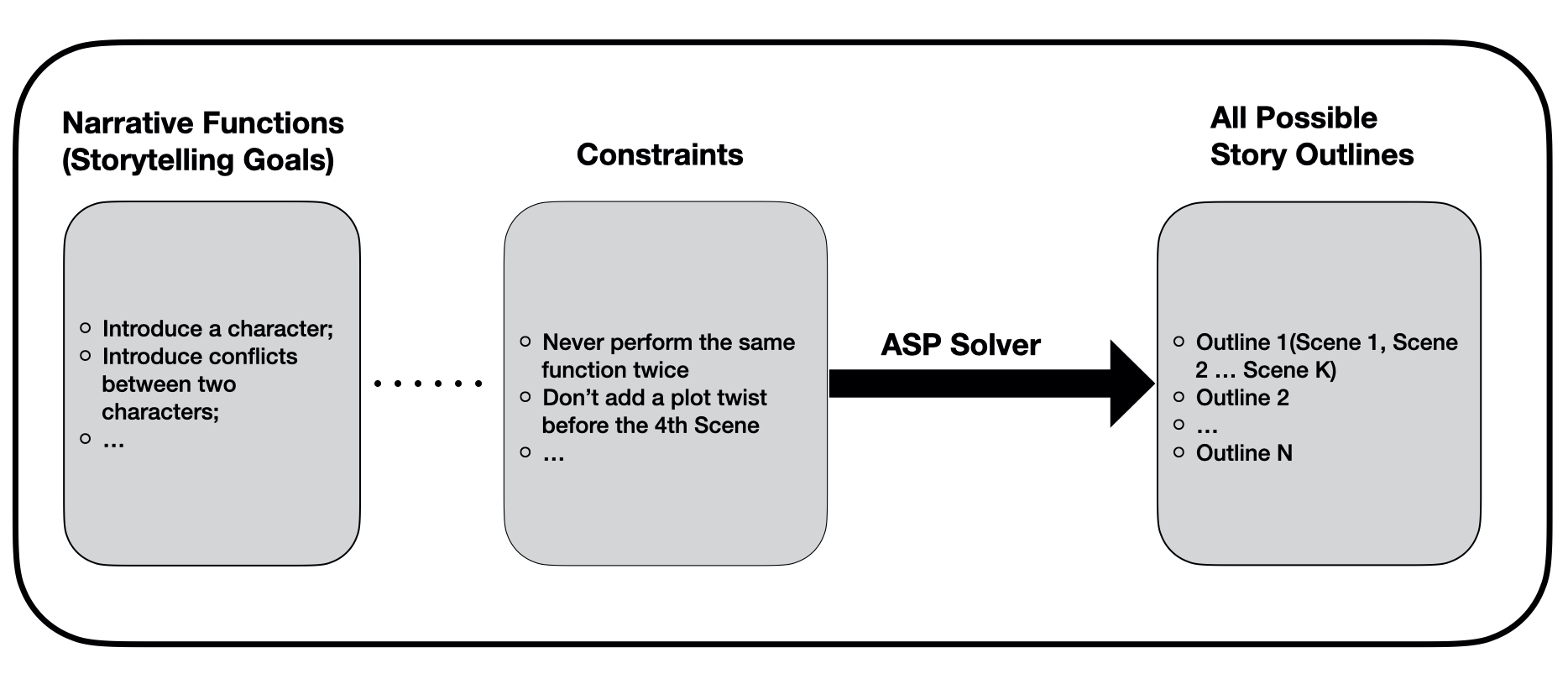}
    \caption{To generate diverse story outlines via ASP, we first specify a set of high-level \emph{narrative functions} or \emph{storytelling goals} that each scene in a story might fulfill. We then define a set of \emph{constraints} that restrict how these narrative functions are allowed to be sequenced and combined. Finally, using an answer set solver, we solve for the set of all valid story outlines that can be produced given the specified narrative functions and constraints.}
    \label{fig:OutlineGeneration}
\end{figure*}

\begin{figure*}
    \centering
    \includegraphics[width=\linewidth]{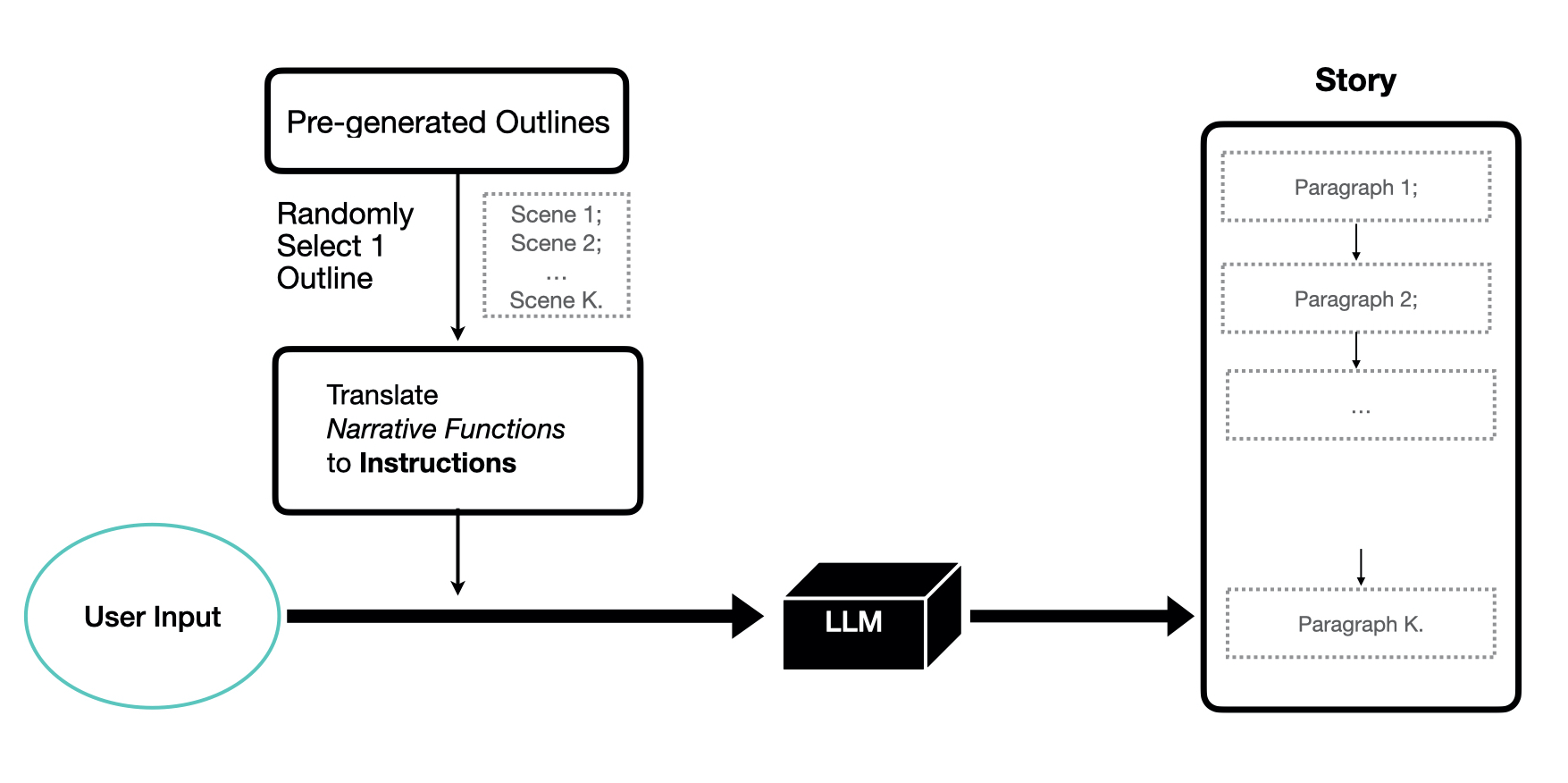}
    \caption{Our overall approach to ASP-guided story generation first takes in a \emph{premise prompt}---a brief textual description of the high-level premise of the story to be generated---and a randomly selected outline from a set of valid outlines generated via ASP. We then translate the sequence of narrative functions defined by the outline to a sequence of \emph{story continuation prompts}. These prompts are fed sequentially into the LLM to generate a sequence of paragraphs, each one taking into account the high-level story premise; the text of the story so far; and the narrative function that this scene (i.e., paragraph) must fulfill. The baseline (unguided) approach is identical, but uses a \emph{neutral} story continuation prompt for each paragraph instead of taking a pre-generated story outline into account.}
    \label{fig:StoryGeneration}
\end{figure*}

Broadly speaking, we generate stories via a two-step process. First, we \textbf{plan} a large and diverse set of potential story outlines by generating the full set of solutions to a small handwritten answer set program that encodes many viable sequences of storytelling goals (Fig. \ref{fig:OutlineGeneration}). Second, we choose an outline at random from this set and \textbf{write} a complete textual story that follows this outline via LLM, conditioning generation on a high-level story premise defined by a short string of text (Fig. \ref{fig:StoryGeneration}).

\subsection{Plan: Outline Pre-Generation with ASP}

In the \textbf{plan} phase, we make use of answer set programming (ASP) to generate diverse story outlines. ASP is a form of constraint solving that employs an \emph{answer set solver} to generate all \emph{stable models}~\cite{StableModelSemantics}, or sets of facts, that satisfy a set of (generally hand-written) logical constraints~\cite{ASPGlance}.

ASP is widely used in procedural content generation research to generate artifacts ranging from story outlines to game levels~\cite{ASPforPCG}. Generally speaking, ASP-for-PCG approaches break down a generative answer set program into two components; the first part defines the basic \emph{elements} that can be combined to form valid artifacts, and the second part defines \emph{constraints} that restrict how these elements are allowed to be combined. In our case, we model story outlines as consisting of a sequence of \emph{scenes}, each of which describes what happens at a certain time point in the story, and each of which fulfills some \emph{storytelling goal} or \emph{narrative function} related to what the story's author is trying to convey.



\paragraph{Timeline and Scenes} We divide the story timeline into a sequence of scenes, for example, seven scenes, which we found to be long enough to create an interesting narrative. In ASP, this is written as:

\begin{verbatim}
#const num_scenes = 7.
scene(1..num_scenes).
\end{verbatim}

Our current implementation ultimately translates each scene to a single paragraph of story text, but scenes could also be rendered as shorter or longer spans of text; as sequences of panels in a comic; as clips of video making up a movie; and so on.

\paragraph{Narrative Functions} Narrative functions are the atomic units of story outlines, representing what storytelling goal the system is trying to accomplish in each scene. In our current implementation, each scene performs exactly one narrative function:

\begin{verbatim}
{ scene_performs_func(S,F)
  : function(F) } = 1 :- scene(S).
\end{verbatim}

So far we have used a relatively minimal set of example narrative functions to test our approach:

\begin{verbatim}
function(intro_char;intro_rival_char;
  add_conflict;add_bonding;
  add_obstacle_towards_major_goal;
  add_obstacle;add_breakthrough;
  add_twist;level_up_obstacle).
\end{verbatim}

\noindent However, we do not intend for this ontology of storytelling actions to be treated as canonical, and we hope that it is revised or replaced with a more rigorously grounded ontology in the future. Future work, for instance, might introduce functions related to the \emph{tone} or \emph{mood} of narration, which could be assigned to scenes in parallel with the plot development functions we have already introduced.


\paragraph{Function Parameters} Some narrative functions can take additional \emph{parameters} that help to further specify a high-level storytelling goal (e.g., introducing a character). In our example outline generator, we use both \emph{personality parameters} to add diversity to character introductions and \emph{obstacle parameters} to add diversity to introduced obstacles. In our experience, these parameters are best introduced when the LLM tends to always satisfy a particular narrative function in very similar ways unless given further guidance. 



\paragraph{Constraints} To prevent the generation of incoherent outlines, we use constraints to rule out certain ways of combining narrative functions. The expressiveness of ASP allows these constraints to be concisely defined. For instance, this constraint prohibits generation of outlines in which two scenes in a row perform the same narrative function:

\begin{verbatim}
:- scene_performs_func(S1,F),
   scene_performs_func(S2,F),
   S2 = S1 + 1.
\end{verbatim}

\noindent And this constraint ensures that we never attempt to add conflict between characters until at least two characters have been introduced:

\begin{verbatim}
:- scene_performs_func(S2,add_conflict),
   0 { scene_performs_func(S1,intro_char)
       : scene(S1), S1 < S2 } 1.
\end{verbatim}

\noindent In general, constraints are used to specify invalid \emph{partial orderings} of narrative functions.

To maximize the diversity of generated outlines, we intentionally specify only a minimal set of constraints. Our example outline generator makes use of just fifteen total constraints, each consisting of no more than three lines of ASP code. Because story outlines are ultimately translated into complete stories by an LLM, ASP constraints do not necessarily have to prevent every problem that might arise in narrative function sequencing: they only need to prevent problems that the LLM is unable to acceptably ``write around'', which can be identified through iterative testing and refinement of the full ASP-to-LLM story generation pipeline.

Though we do not do so here, constraints could also be used to restrict outline generation according to user preferences. For example, a constraint might be used to force generation of outlines that contain at least a certain amount of bonding between characters. An interactive story generation tool based on our approach might even provide a user interface for enabling and disabling various optional constraints.


\paragraph{Implementation} We implement ASP functionality using the \texttt{clingo} library~\cite{Potassco}. Having defined an answer set program that describes the design space of valid story outlines, we run the answer set solver to generate all possible outlines that satisfy the given rules. We then save these outlines to a file for later use. Our initial evaluation involved the pregeneration of a set of 401,990 distinct outlines. 

\subsection{Write: Story Generation with LLM}

In the \textbf{write} phase, we use an LLM to expand an outline generated via ASP to a complete written story. This phase takes two key inputs:

\begin{enumerate}
    \item User input: a short textual \emph{premise prompt} that describes the high-level premise of the story to be generated.
    \item Outline: a randomly selected story outline from the pre-generated outline pool. In our current implementation, an outline consists of a sequence of seven narrative functions, one function for each scene. Each scene will be written out as one paragraph of story. An example outline looks like this:

    \texttt{[`introduce\_character:sunny', `add\_obstacle:supernatural', `introduce\_character:mysterious', `add\_twist', `introduce\_character:clumsy', `introduce\_rival\_character', `level\_up\_obstacle']}

\end{enumerate}

\subsubsection{Translate Narrative Functions to Writing Instructions} 

We use a dictionary to map each narrative function in the chosen outline to a detailed, human-authored \emph{writing instruction} for the LLM to follow.

For example, for the narrative function \texttt{introduce\_character:sunny}, the resulting instruction is: \textit{``introduce an optimistic and brave character who would fit organically. Consider how this character's presence can be connected to the current plot, setting, or other characters. Craft a scene or situation that allows for a smooth integration of this character, ensuring that their introduction doesn't feel forced or disconnected from the narrative. Provide context for their sunny and brave nature through their actions, dialogue, or the reactions of other characters, while maintaining the flow and consistency of the story. This character's introduction should contribute to the development of the plot or the growth of other characters in a meaningful way, while maintaining the flow and consistency of the story''}.

\subsubsection{Generate Story Using Prompts}
We use the \texttt{gpt-3.5-turbo} LLM, via the OpenAI API, to render a selected outline into a full story. A consistent system prompt is used to set the context for the LLM, instructing it to act as a fiction writer.

For the first paragraph of the generated story, the model is prompted with the story premise as well as the appropriate writing instruction for the first narrative function in the outline: \emph{``You're writing a story about: \{premise\}. Write the first paragraph of the story. In this paragraph, \{instruction\}.''} The resulting paragraph is then appended to the conversation history.

For subsequent paragraphs, the same process is repeated, using a slightly modified story continuation prompt: \emph{``Write the next paragraph of the story. Remember the story is about: \{premise\}. In this paragraph, \{instruction\}.''} The complete conversation history is fed into subsequent paragraph generation requests as additional context, and the newly generated paragraph is then appended to the conversation history. This process is repeated until every entry in the story outline has been expanded to a full paragraph of text.

\subsection{Baseline: Unguided Generation}
In addition to an ASP-guided LLM-based story generator, we also implement an unguided LLM-based generator as a baseline for evaluation. This baseline generator follows a nearly identical process to that described above, with one key difference: the paragraph generation prompts issued to the baseline generator do not include the translated instructions from the outline. Instead, they rely solely on the user-supplied premise to guide  story generation.

\section{Evaluating Story Diversity}

\subsection{Procedure}

\begin{figure*}
    \centering
    \includegraphics[width=\linewidth]{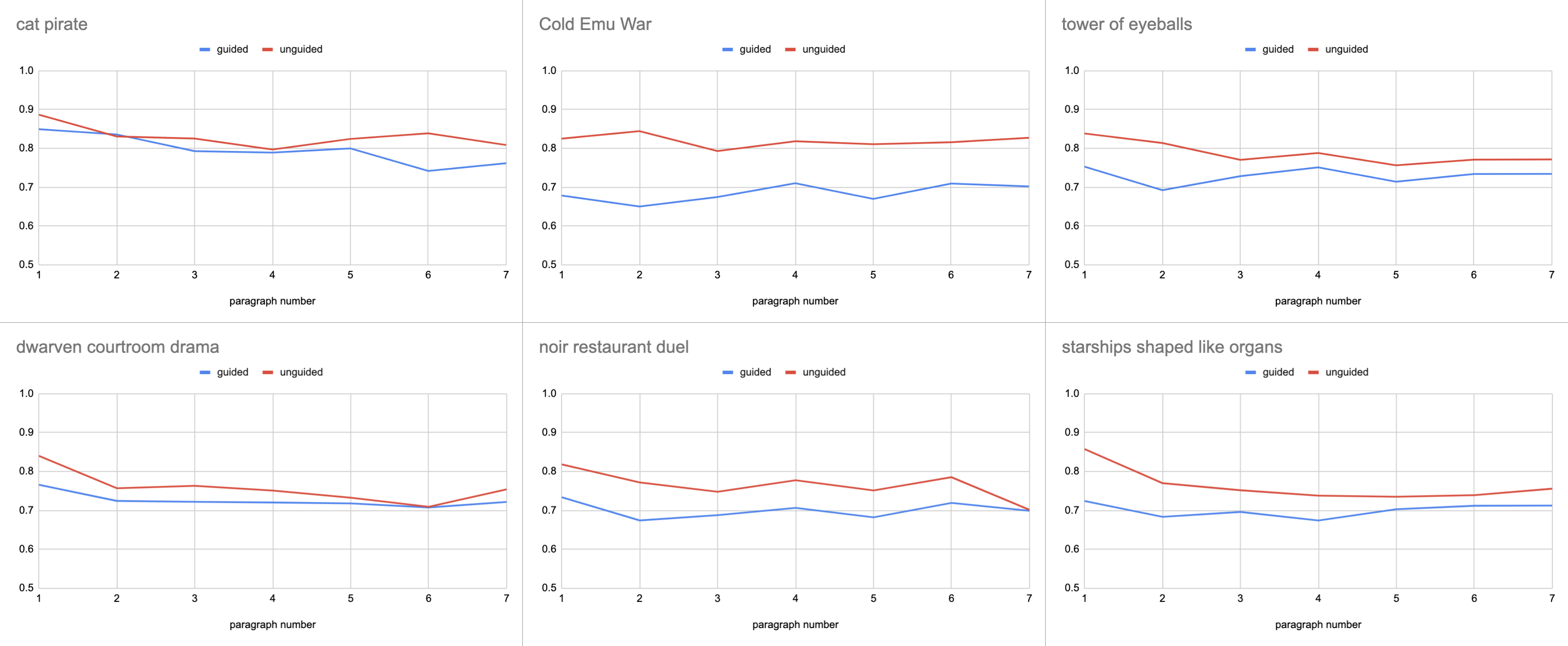}
    \caption{Comparative paragraph-by-paragraph homogeneity scores for stories generated around six different premises, using both an ASP-guided (\textcolor{blue}{blue}) and an unguided (\textcolor{red}{red}) approach. For each premise, we generated a set of ten seven-paragraph stories using each approach, then visualized how semantically homogenous these story sets were at each paragraph. Lower homogeneity scores are better; the maximum possible score is 1 (all stories in the set are identical) and the minimum possible score is 0 (stories share nothing in common whatsoever).}
    \label{fig:Homogeneity}
\end{figure*}

We evaluate our approach by generating parallel sets of ten stories each using our (ASP-guided) approach and a na\"ive (unguided) baseline, then comparing the semantic homogeneity of each story set via a sentence embedding model. Intuitively, we can calculate a \emph{homogeneity score} for a set of texts via the following procedure:

\begin{enumerate}
    \item Embed each text.
    \item Average these embeddings together to calculate a \emph{central} embedding for the set as a whole.
    \item Calculate the cosine similarity between this central embedding and the embedding of each text in the set.
    \item Average together these similarity scores to determine a summary homogeneity score for the set as a whole.
\end{enumerate}

\noindent This is similar to the approach taken in several past studies of the homogenization effects of AI-based creativity support tools~\cite{PadmakumarHe,HomogenizationEffects}.

Homogeneity scores resulting from this approach range from 0 to 1, with higher scores representing more homogenous (i.e., less diverse) sets of texts. A score of 1 corresponds to a set of identical texts, while a score of 0 corresponds to a set of texts that share nothing at all in common.

In practice, an entire generated story is too long to compress into a single embedding using a standard sentence embedding model, so for our evaluation we instead represent each story as a \emph{sequence} of embeddings: one embedding for each paragraph of generated text. We then calculate a separate homogeneity score for each \emph{paragraph index} in a set of stories that each contain the same number of paragraphs. A set of stories that all begin with the same opening paragraph but diverge substantially in the second paragraph, for instance, will have a very high homogeneity score for the first paragraph but a lower homogeneity score for the second.

To validate that our approach improves story diversity across a wide range of different story premises, we repeat our evaluation procedure with six different \emph{premise prompts}: short text strings that a user might input to a story generator to describe the basic premise of a story (e.g., ``cat pirate'', ``dwarven courtroom drama''). 

We use the pretrained \texttt{all-MiniLM-L6-v2} sentence embedding model to conduct semantic similarity calculations. This model has proven to align relatively well with human judgments of semantic similarity in a divergent ideation context~\cite{HomogenizationEffects}, and the general class of transformer-based sentence embedding models to which this model belongs has been shown to outperform word embedding models that were previously evaluated as the gold standard for semantic similarity calculation in creativity research~\cite{SemDis,Dumas}. \texttt{all-MiniLM-L6-v2} is available as part of the open source SentenceTransformers library~\cite{SentenceTransformers}, supporting replication of our evaluation approach.

\subsection{Results}

In general, we found that our ASP-guided approach consistently generated more semantically diverse sets of stories than an unguided approach, regardless of premise and paragraph number (Fig. \ref{fig:Homogeneity}). 

Some premises benefited more from ASP guidance than others; for instance, stories generated in response to the ``Cold Emu War'' premise were much more diverse when ASP guidance was employed. Although the reason for these premise-level differences are not yet clear to us, they seem consistent across multiple different evaluation runs: for instance, story sets produced in response to the ``cat pirate'' premise consistently seem to benefit the least from ASP guidance in terms of diversity. Perhaps this is because, during pretraining, the LLM was exposed to more closely relevant training data for some of our evaluated story premises than for others; this could give the LLM an especially strong prior for what stories that respond to certain premises ``should'' look like, resulting in ASP guidance exerting less of an effect on overall story direction for these premises.

We also observed that earlier paragraphs tended to be more homogenous than later paragraphs regardless of whether ASP guidance was used. However, this trend was relatively weak and did not hold across all premises. Insofar as later paragraphs \emph{are} more diverse than earlier paragraphs, this may be because later paragraphs are generated in the context of more preceding text, so minor differences in earlier paragraphs may ``pile up'' and be reinforced in later paragraphs as the later paragraphs become progressively more conditioned on earlier story contents.

\section{Limitations and Future Work}
\paragraph{Evaluation limitations} Human readers are the ultimate arbiters of story quality, and although we believe stories generated with ASP guidance are at least no worse than unguided stories, we have not yet conducted any systematic human subjects evaluation of our approach. We use sentence embedding to evaluate the diversity of generated stories; however, embedding models do not align completely with human judgments of semantic similarity~\cite{HomogenizationEffects}, and seem to over-prioritize some aspects of sentences (e.g., nouns) compared to others~\cite{SentenceTransformerBias}. The replicability of our evaluation is also limited by the use of ChatGPT 3.5 (a closed-source, closed-data, and closed-weights commercial model) as our target LLM, though we would expect to see similar results with any comparable instruction-tuned LLM.

\paragraph{Controllability limitations} ASP guidance improves the diversity of generated stories by adding (constrained) \emph{randomness} to the story generation process. This reduces the proportional contribution of language model defaults to the output stories, but doesn't increase the proportional contribution of \emph{user intent}; in fact, adding randomness without adding more user-controlled input information \emph{reduces} the overall contribution of user intent to the output. This may exacerbate the ``dearth of the author'' effect, by which creative artifacts produced using AI-based tools can fail to strongly reflect the user's authorial intent~\cite{DearthOfTheAuthor}.

\paragraph{Future work} In the future we plan to build an approachable tool that lets users interactively constrain the outline generation (ASP) portion of our generative pipeline, perhaps drawing interaction design inspiration from existing constraint-based tools for mixed-initiative storytelling~\cite{Mimisbrunnur,LooseEnds} and game design~\cite{Germinate}. The existence of such an interface would bring the injected randomness back under user control to some extent, helping to reverse the dilution of user intent. We also plan to investigate the use of an LLM to generate new ASP constraints in response to open-ended natural-language statements of storytelling intent from users, in much the same way that \citeauthor{ThereAndBackAgain} use an LLM to generate narrative planning domains from natural-language descriptions of a high-level story premise. This could enable an even greater degree of user control over story outline generation.

\section{Related Work}
In the procedural content generation research community, ASP is widely used to specify design spaces of creative artifacts~\cite{ASPforPCG}. We are aware of two prior approaches to generating story outlines with ASP: one that leverages an ASP encoding of event calculus~\cite{RoleModel} and one that encodes a traditional narrative planner as an answer set program~\cite{MartensASP}. Both of these approaches differ from ours in that they use ASP to generate sequences of \emph{plot events}; consequently, the resulting answer set programs are relatively complicated due to the need to explicitly encode every action that a character might potentially take, including associated preconditions and postconditions. Our approach, on the other hand, uses ASP to generate sequences of \emph{storytelling goals}; fewer constraints are needed to specify valid outlines in this looser formalism, resulting in a much simpler program.

Luminate~\cite{Luminate} introduces a new purely LLM-based plan-and-write approach that explicitly aims to generate diverse sets of stories. Given an initial premise, the system first uses an LLM to generate \emph{dimensions} of a design space for stories that address this premise; then uses the LLM again to generate possible \emph{values} for those dimensions; then uses it a third time to generate a structured set of stories that combine these values in diverse ways. This bears some resemblance to how we explicitly construct a design space of story outlines with ASP, and is in fact more flexible than our approach, but does not admit the specification of constraints that \emph{disallow} certain combinations of values (which ASP readily permits).


Outside of plan-and-write (i.e., prompting-focused) approaches, several attempts have also been made to improve the diversity of LLM-generated text via more general, model-focused approaches. These approaches include the use of diverse decoding methods to extract more diverse outputs from a fixed LLM~\cite{DiverseDecoding}; quality-diversity approaches such as QDAIF~\cite{QDAIF}; and finetuning approaches to force the generation of ``diffuse distribtions'' in contexts where output diversity is desirable~\cite{DiffuseDistributions}. All of these approaches are promising in their own right for improving output diversity, though they do not tend to provide new control affordances for story generation in the same way as plan-and-write approaches.

\section{Conclusion}
We have demonstrated the use of answer set programming (ASP) to guide LLM-based story generation toward greater diversity, using a concise answer set program (more flexible than a full-fledged narrative planning domain) to define a generator of valid story outlines and an LLM to expand these outlines into complete stories. We believe this approach represents a new reference point in the design space of neurosymbolic approaches to story generation, and a potentially compelling way to combine the open-domain story generation capabilities of LLMs with the greater content planning abilities of rules-based methods. We make our code available as a jumping-off point for future research in this direction.\footnote{\url{https://github.com/mkremins/spleenwort}}

\section*{Acknowledgments}
This work was supported by University Research Grant GR102947, ``A Neurosymbolic Approach to Creativity Support for Narrative Writing'', from Santa Clara University.

\bibliography{custom}


\end{document}